\documentclass[letterpaper]{article} % DO NOT CHANGE THIS
\usepackage[draft]{aaai25}  % DO NOT CHANGE THIS
\usepackage{times}  % DO NOT CHANGE THIS
\usepackage{helvet}  % DO NOT CHANGE THIS
\usepackage{courier}  % DO NOT CHANGE THIS
\usepackage[hyphens]{url}  % DO NOT CHANGE THIS
\usepackage{graphicx} % DO NOT CHANGE THIS
\usepackage{natbib}  % DO NOT CHANGE THIS AND DO NOT ADD ANY OPTIONS TO IT
\usepackage{caption} % DO NOT CHANGE THIS AND DO NOT ADD ANY OPTIONS TO IT
\usepackage{amsfonts}
\usepackage{bm}
\usepackage{upgreek}
\usepackage{amsmath}
\DeclareMathOperator*{\argmin}{arg\,min}
\usepackage{dsfont}
\usepackage{breqn}
\usepackage{color}
\usepackage[inline]{enumitem}
\usepackage{etoolbox}

\usepackage{algorithm}
\usepackage{algorithmic}

\usepackage{newfloat}
\usepackage{listings}
\DeclareCaptionStyle{ruled}{labelfont=normalfont,labelsep=colon,strut=off} % DO NOT CHANGE THIS
\floatstyle{ruled}
\newfloat{listing}{tb}{lst}{}
\floatname{listing}{Listing}
\title{Angle of Arrival Estimation with Transformer: \\ A Sparse and Gridless Method with Zero-Shot Capability}
\author{
    Zhaoxuan Zhu, Chulong Chen and Bo Yang
}
\affiliations{
    Motional, Inc. \\
    \{zhaoxuan.zhu, chulong.chen, bo.yang\}@motional.com
}

\usepackage{bibentry}
\begin{document}

\maketitle

\begin{figure*}[htb]
\centering
\includegraphics[width=0.99\textwidth]{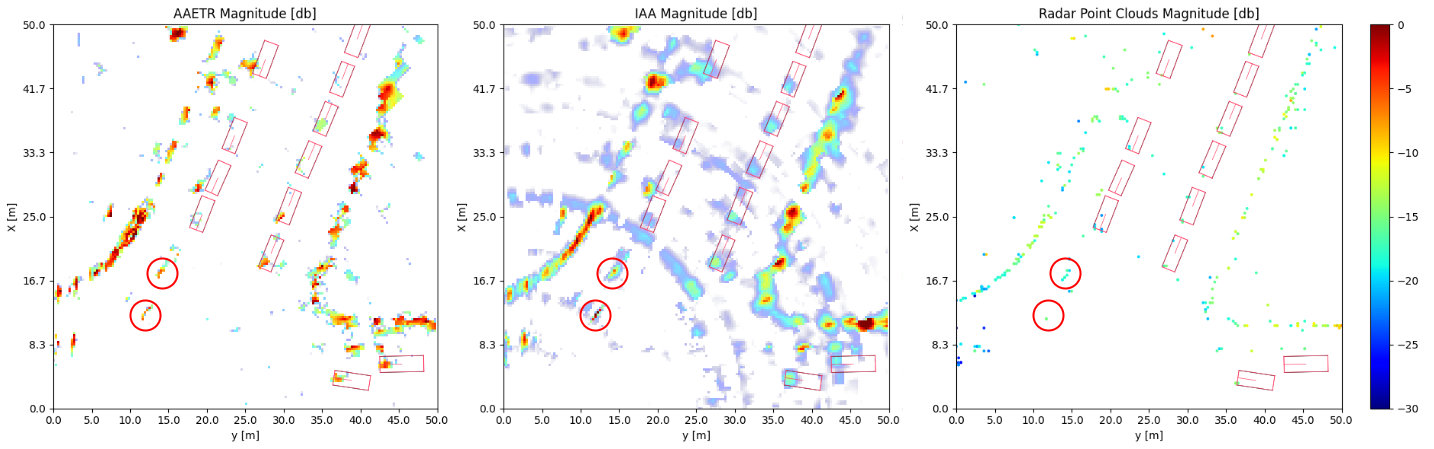} 
\caption{Zoom-in views of the BEV feature maps from AAETR, IAA and the radar detections by embedded DSP.
AAETR demonstrates significantly lower FP detections and improved side-lobe suppression compared to the classic IAA method in the scene captured by the camera view in Figure \ref{fig: front left camera view}.
Both AAETR and IAA provide much richer semantic information for downstream perception tasks than the sparse point clouds from on-device radars.
For instance, the radar images from AAETR and IAA accurately capture vehicle dimensions in the scene and clearly detect thin street poles (highlighted by red circles).
These low-level representations preserve valuable semantic information such as size and shape for subsequent perception tasks.
} 
\label{fig: front left radar view}
\end{figure*}

\begin{abstract}
Automotive Multiple-Input Multiple-Output (MIMO) radars have gained significant traction in Advanced Driver Assistance Systems (ADAS) and Autonomous Vehicles (AV) due to their cost-effectiveness, resilience to challenging operating conditions, and extended detection range.
To fully leverage the advantages of MIMO radars, it is crucial to develop an Angle of Arrival (AoA) algorithm that delivers high performance with reasonable computational workload.
This work introduces AAETR (\textbf{A}ngle of \textbf{A}rrival \textbf{E}stimation with \textbf{TR}ansformer) for high performance gridless AOA estimation.
Comprehensive evaluations across various signal-to-noise ratios (SNRs) and multi-target scenarios demonstrate AAETR's superior performance compared to super resolution AOA algorithms such as Iterative Adaptive Approach (IAA) \cite{yardibi2010source}. 
The proposed architecture features efficient, scalable, sparse and gridless angle-finding capability, overcoming the issues of high computational cost and straddling loss in SNR associated with grid-based IAA.
AAETR requires fewer tunable hyper-parameters and is end-to-end trainable in a deep learning radar perception pipeline.
When trained on large-scale simulated datasets then evaluated on real dataset, AAETR exhibits remarkable zero-shot sim-to-real transferability and emergent sidelobe suppression capability. This highlights the effectiveness of the proposed approach and its potential as a drop-in module in practical systems.
\end{abstract}

\begin{figure}[htb]
\centering
\includegraphics[width=0.99\columnwidth]{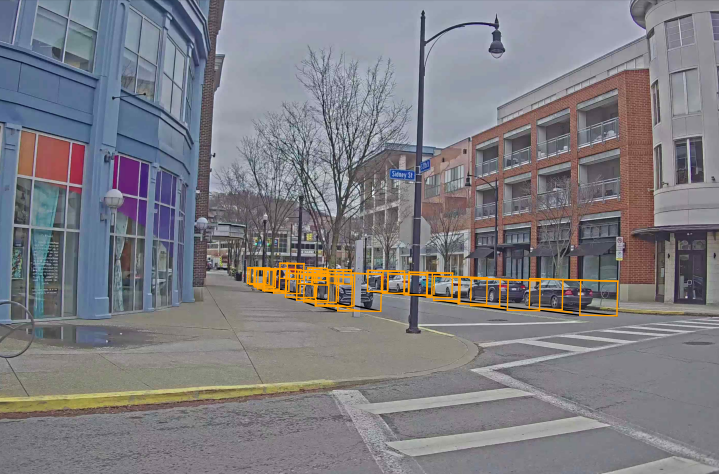}
\caption{The camera view of the front left side of the ego vehicle in the scene in Figure \ref{fig: sample rendering 1}.}
\label{fig: front left camera view}
\end{figure}
% Uncomment the following to link to your code, datasets, an extended version or similar.
%
% \begin{links}
%     \link{Code}{https://aaai.org/example/code}
%     \link{Datasets}{https://aaai.org/example/datasets}
%     \link{Extended version}{https://aaai.org/example/extended-version}
% \end{links}
\section{Introduction}
As the automotive industry progresses towards commercialization of Advanced Driver Assistance Systems (ADAS) and Autonomous Vehicles (AV), there is an increasing adoption of Multiple-input Multiple-output (MIMO) radar technology.
Compared to lidar systems, radars offer significant advantages in terms of cost-effectiveness, resilience to adverse weather conditions, and extended detection range \cite{patole2017automotive, engels2017advances}.
However, the conventional radar data representation, primarily point clouds extracted through Fast Fourier Transform (FFT) and Constant False Alarm Rate (CFAR) processing \cite{richards2005fundamentals}, presents challenges.
These point clouds often suffer from sparsity and a lack of semantic information, which ultimately hinders the performance of crucial downstream tasks such as object detection and segmentation.
This limitation underscores the need for more sophisticated radar signal processing and data representation techniques to fully leverage the potential of MIMO radar systems in autonomous driving applications.

In contrast to point cloud representations, several low-level radar data formats are available at various stages of the radar signal processing chain.
These include Analog-to-Digital Converter (ADC) data, Range-Doppler (RD) cubes, and Range-Angle-Doppler (RAD) cubes or radar images.
While these low-level representations require more sophisticated processing algorithms and incur higher storage costs \cite{rebut2022raw}, they ultimately yield superior performance in downstream perception tasks. 
This improved performance is evident whether radar is used as the sole sensing modality \cite{palffy2020cnn,yang2023adcnet} or in fusion with camera data \cite{liu2024echoes, hwang2022cramnet}.

Within the radar signal processing chain, the Range-Doppler (RD) cube undergoes Direction of Arrival (DOA), or interchangeably in this work Angle of Arrival (AOA), to generate the Range-Angle-Doppler (RAD) cube.
The efficacy of the AOA algorithm directly influences the quality and resolution of the resulting RAD cube, which can subsequently be projected into a Bird's-Eye-View (BEV) representation for downstream tasks  \cite{major2019vehicle, zheng2023time}.
While extensive research has been conducted on AOA estimation using conventional approaches, the practical application of these methods is often constrained by either high computational demands \cite{yardibi2010source, liao2016music} or the necessity for multiple temporal snapshots \cite{schmidt1982signal}. 
These limitations underscore the need for more efficient and adaptable AOA estimation techniques that can meet the real-time processing requirements of automotive radar systems while maintaining high accuracy.

In recent years, deep learning-based AOA algorithms have emerged in the literature \cite{papageorgiou2021deep, zheng2023deep},  offering faster inference speeds while achieving performance comparable to conventional approaches.
However, existing studies have primarily focused on two distinct areas: enhancing end-to-end perception tasks such as 3D object detection and segmentation, or improving DOA estimation performance under limited and synthetic scenarios.
% This dichotomy in research focus has left a gap in comprehensive evaluation of deep learning-based AOA methods across a wide range of realistic conditions, particularly in the context of automotive radar applications. 
% There remains a need for a thorough assessment of these algorithms' performance, robustness, and generalization capabilities in real-world automotive sensing environments.
The dichotomy in research focus has created a gap in evaluating deep learning-based AOA methods for automotive radar, particularly regarding their real-world performance and robustness.

In this work, we introduce AAETR (\textbf{A}ngle of \textbf{A}rrival \textbf{E}stimation with \textbf{TR}ansformer), a novel approach inspired by the seminal object detection framework DETR \cite{carion2020end}.
Our model architecture comprises an encoder that processes array signals and a decoder that predicts targets using cross-attention mechanisms between encoded messages and learnable queries.
Leveraging the powerful capabilities of transformer-based models and training on an extensive, efficiently generated yet realistic simulated dataset, AAETR demonstrates substantial improvements in AOA performance across comprehensive evaluations. 
Notably, AAETR exhibits excellent zero-shot generalization capabilities when applied to data collected from real-world driving scenarios, as illustrated in in Figure \ref{fig: front left radar view} and Figure \ref{fig: front left camera view}.
This remarkable transfer from simulated to real-world data underscores the robustness and adaptability of our proposed method in practical automotive radar applications.

The key contributions of this work are threefold.
\begin{enumerate*}[label={(\arabic*)}]
    \item We introduce a fully differentiable, transformer-based AOA model that significantly outperforms well-established super-resolution methods. The model's sparse detection architecture enables gridless estimation, offering a substantial computational advantage over traditional methods, especially at high grid resolutions.
    \item Our proposed model demonstrates remarkable zero-shot transferability from synthetic datasets to real-world radar data, highlighting its robustness and adaptability to practical applications.
    \item We present a novel evaluation methodology that correlates downstream object detection and segmentation task performance with spectrum estimation accuracy, providing a more comprehensive assessment of the model's practical utility.
\end{enumerate*}
% These contributions collectively advance the state-of-the-art in AOA estimation for automotive radar systems, offering both theoretical improvements and practical advantages for real-world deployment.

\section{Related Works}

\begin{figure*}[htb]
\centering
\includegraphics[width=0.99\textwidth]{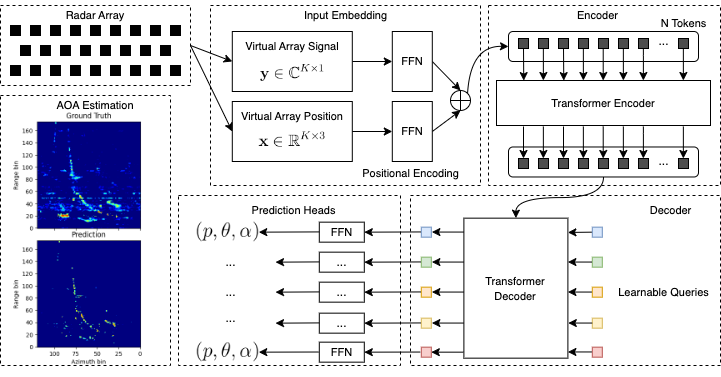} 
\caption{The overall network architecture. The radar signal and the array positions are fed into the transformer encoder as input. The encoded messages are then attended by the learnable queries in the decoder. The decoded messages are then mapped into the regression parameters and detection logits with FFN.}
\label{fig: aaetr overall architecture}
\end{figure*}

\subsection{Angle of Arrival Algorithms}
One of the pioneering AOA algorithms is MUltiple SIgnal Classification (MUSIC) \cite{schmidt1982signal}.
MUSIC is a grid-based method that requires multiple snapshots for accurate estimation, especially under low Signal-to-Noise Ratio (SNR) conditions.
Root-MUltiple SIgnal Classification (R-MUSIC) \cite{barabell1983improving} is a grid-less method that achieves improved accuracy.
Single snapshot MUSIC is proposed in \cite{liao2016music}, which is more suitable for automotive applications \cite{sun2020mimo}. 
The Iterative Adaptive Approach (IAA) \cite{yardibi2010source} is another grid-based, iterative and non-parametric method that operates with a single snapshot. 
A sparsity-promoting approach with hard and soft thresholding was later proposed in \cite{rowe2013sparse}.
The main challenge associated with iterative approaches is their high computational complexity, which can hinder their application in real-time use cases.

Recently, deep learning based approaches have also been explored. 
A grid-based method is developed with Convolutional Neural Network (CNN) in \cite{papageorgiou2021deep}. Despite its robustness under low SNR, it requires multiple snapshots.
Inspired by IAA, Unrolling IAA (UAA) with recurrent neural network is proposed for efficient inference \cite{zheng2023deep}. 
In \cite{wang2024human}, a 3D CNN based network is developed for ADC to RAD cube signal processing for human detection in their in-house through-wall dataset.

\subsection{Low Level Radar for Perception}
In the recent years, low level radar representations have been extensively studied to ADAS and AV perception systems either as the only modality or fused with other sensing technologies. 
A graph convolution network is developed to leverage the low level radar data for long range 3D object detection under complex scenarios \cite{meyer2021graph}.
MVDNet \cite{qian2021robust} fuses imaging radar and lidar point clouds and demonstrates the superior 3D object detection performance in long range and foggy driving conditions compared to systems with single modality.
CramNet \cite{hwang2022cramnet} uses cross attention to fuse the radar Range Azimuth image with camera images for 3D object detection. 

\section{AAETR} \label{section: aaetr}
In this section, the spectral estimation problem is formulated as a detection problem.
The input to the algorithm is the measured data vector $\textbf{y} \in \mathbb{C}^{K \times 1}$, and the virtual array element positions $\textbf{x} \in \mathbb{R}^{K \times 3}$ is also available, where $K$ is the number of virtual array elements.
The output of the algorithm are the predicted angle $\tilde{\boldsymbol\uptheta} \in \mathbb{R}^{M \times 1}$, magnitude $\tilde{\boldsymbol\upalpha} \in \mathbb{R}^{M \times 1}$ and confidence of $\tilde{\textbf{p}} \in \mathbb{R}^{M \times 1}$ of the detection targets, where $M$ is the maximal number of detection targets as a hyperparameter to the algorithm.

For the detection problem, a transformer model inspired by DETR \cite{carion2020end} is proposed as illustrated in Figure \ref{fig: aaetr overall architecture}.
The real and imaginary parts of the input measured data vector $\textbf{y} \in \mathbb{C}^{K \times 1}$ and the virtual array element positions $\textbf{x} \in \mathbb{R}^{K \times 3}$ are first projected to high dimensional embedding space $\bar{\textbf{y}} \in \mathbb{R}^{K \times D}$ and $\bar{\textbf{x}} \in \mathbb{R}^{K \times D}$ and subsequently fed into 6 transformer encoder blocks as the input and the positional encoding, respectively.
Each transformer encoder block consists of a layer normalization module (LN), a multi-head self-attention module and the feed-forward networks (FFNs).     
The encoded messages $\bar{\textbf{z}} \in \mathbb{R}^{K \times D}$ and the positional encoding  $\bar{\textbf{x}}$ are used as the keys and the values in the transformer decoder blocks.
The queries $\bar{\textbf{q}} \in \mathbb{R}^{M \times 1}$, along with the positional encoding for the queries, are learnable and initialized randomly.
Each transformer decoder consists of a LN module, a multi-head cross-attention module and the FNNs.
The last layer of the FNNs maps the learnable queries into the output space $\tilde{\boldsymbol\uptheta} \in \mathbb{R}^{M \times 1}$ $\tilde{\boldsymbol\upalpha} \in \mathbb{R}^{M \times 1}$ and $\tilde{\textbf{p}} \in \mathbb{R}^{M \times 1}$. 

A modified set prediction loss in \cite{carion2020end} is used for training.
Given the $N$ ground truth targets, the ground truth angles $\boldsymbol\uptheta \in \mathbb{R}^{M \times 1}$ and magnitudes $\boldsymbol\upalpha \in \mathbb{R}^{M \times 1}$ are padded with $M - N$s no detection. 
We use $c_i =  \oslash$ to indicate the ith ground truth is a padded detection.  
The bipartite matching between the ground truth and prediction sets is performed to find the optimal permutation of $M$ elements $\sigma \in P$ as follows,
\begin{equation}
    \sigma^* = \argmin_{\sigma \in \mathcal{P}} \sum_{i}^{M} \mathcal{L}_{\text{match}}\left( \theta, \alpha, c_i, \tilde{\theta}_{\sigma(i)}, \tilde{\alpha}_{\sigma(i)}, \tilde{p}_{\sigma(i)} \right),
\end{equation}
where 
\begin{dmath}
     \mathcal{L}_{\text{match}}\left( \theta, \alpha, c_i, \tilde{\theta}_{\sigma(i)}, \tilde{\alpha}_{\sigma(i)}, \tilde{p}_{\sigma(i)} \right)= -\mathds{1}_{\left\lbrace c_i \neq \oslash\right\rbrace} \log \tilde{p}_{\sigma(i)} + w_{\theta} \cdot \mathds{1}_{\left\lbrace c_i \neq \oslash \right\rbrace} \left\lVert \theta_i - \tilde{\theta}_{\sigma(i)} \right\rVert_1 + w_{\alpha} \cdot \mathds{1}_{\left\lbrace c_i \neq \oslash \right\rbrace} \left\lVert \alpha - \tilde{\alpha}_{\sigma(i)} \right\rVert_1,
\end{dmath}
and $w_{\alpha}$ and $w_{\theta}$ are the weights to the regression loss terms associated with magnitude and angle, respectively. 
With the bipartite matching results, the training objective is as follows, 
\begin{dmath}
     \mathcal{L}_{\text{train}}\left( \theta, \alpha, c_i, \tilde{\theta}_{\sigma^*(i)}, \tilde{\alpha}_{\sigma^*(i)}, \tilde{p}_{\sigma^*(i)} \right)= -\mathds{1}_{\left\lbrace c_i \neq \oslash\right\rbrace} \log \tilde{p}_{\sigma^*(i)} + w_{\theta} \cdot \mathds{1}_{\left\lbrace c_i \neq \oslash \right\rbrace} \left\lVert \theta_i - \tilde{\theta}_{\sigma^*(i)} \right\rVert_1 + w_{\alpha} \cdot \mathds{1}_{\left\lbrace c_i \neq \oslash \right\rbrace} \left\lVert \alpha - \tilde{\alpha}_{\sigma^*(i)} \right\rVert_1.
\end{dmath}

\section{Comprehensive Evaluation Method}
Precise multi-target detection and localization are crucial for perception tasks, as demonstrated by lidar-first perception systems. This section introduces a comprehensive evaluation framework for radar AOA estimation algorithms, designed to assess performance in core perception tasks like 3D object detection. The evaluation focuses on three key areas: detection capability using precision-recall metrics, measurement accuracy in angle and magnitude, and algorithm robustness across varying SNRs and target counts. this framework provides a rigorous methodology for assessing angle-finding algorithms under diverse operational conditions relevant to complex, real-world applications.

\subsubsection{Detection Capability: Precision-Recall Metrics}
At the core of many perception tasks is precision-recall analysis, which quantifies a system's detection performance. We adopt this concept to evaluate AOA algorithms in a multi-target setup.
Given $M$ predicted targets $\left(\tilde{\theta}_m, \tilde{\alpha}_m, \tilde{p}_m \right)$ for $m=1, 2, ... M$, the predictions with low probability are first filtered out by confidence thresholds.
For each scenario, true positives (TP) are detections within $\pm 0.5$ degrees of any ground truth target. False positives (FP) are detections not corresponding to any ground truth target within this threshold, while false negatives (FN) are undetected ground truth targets.

\subsubsection{Measurement Error}
We assess AOA estimation accuracy among true positive detections using average L1 errors in angle (degrees) and magnitude (dB) between predictions and ground truths.

\subsubsection{Dynamic Range and Ccenario Configuration}
The methods are evaluated under a challenging 13 dB dynamic range, representing the spread from the weakest to the strongest target signal. We vary target counts across scenarios from 2 to 10 per range-Doppler bin, introducing different complexity levels. Additionally, we perform evaluations across SNR levels ranging from 15 dB to 35 dB.

\section{Experiments}

\subsection{Synthetic Dataset}
In this study, a real long range radar (LRR) equipped on our fleet vehicle is used for simulation, training, evaluation and visualization.
The radar features a sparse MIMO array with 6 transmitters and 8 receivers, resulting in a 48 virtual elements.
To train the transformer model in the supervised fashion, a large amount of high-quality data is required. 
Starting by using IAA as the model to provide ground truth, we encountered two issues.
First, the trained model has a performance upper bound set by IAA, and it inherits the undesirable side lobe issue from IAA.
Second, performing IAA on a large amount of real sensor data is computationally intensive and can slow down the model training extensively. 
To address these issues, we propose to use a physics-based high-fidelity simulation pipeline to generate ground truth data efficiently in a large scale.

For each training sample, the number of targets are first sampled uniformly as $N\sim \mathcal{U}\left\lbrace 0, N_{\text{max}}\right\rbrace$, where $N_{\text{max}}$ is the maximum number of targets. We choose the maximum number of targets in the generation that balance the design of the array and the complexity of the real-world driving scenario. To ensure generalization, the angle $\theta$ and magnitude $\alpha$ of each target are uniformly sampled as $\theta \sim \mathcal{U}\left\lbrace \theta_{\text{min}}, \theta_{\text{max}} \right\rbrace$ and $\alpha \sim \mathcal{U}\left\lbrace \alpha_{\text{min}}, \alpha_{\text{max}} \right\rbrace$.

Consider a radar sensor with \(K\) array elements. When multiple targets are present, the signal received by the \(k\)-th element at time \(t\) can be modeled as the sum of the contributions from each target, represented by \(M\) targets:
\begin{equation}
y_k(t) = \sum_{m=1}^M \alpha_m s(t - \tau_{k,m}(\theta_m)) + n_k(t)
\end{equation}
where \(y_k(t)\) is the signal received at the \(k\)-th element; \(\alpha_m\) is the amplitude of the \(m\)-th target; \(\tau_{k,m}(\theta_m)\) is the delay of the signal from the \(m\)-th target at the \(k\)-th element; $n_k(t)$ is the noise and interference at the \(k\)-th element, \(s(t)\) the transmitted signal. 
The time delay at each element is represented as a steering vector for the \(m\)-th target, \(\mathbf{a}_m(\theta_m)\) given by 
\begin{equation}
\mathbf{a}_m(\theta_m) = \alpha_m \begin{bmatrix}
    e^{-j\omega \tau_{1,m}(\theta_m)}\\ e^{-j\omega \tau_{2,m}(\theta_m)}\\ \ldots \\ e^{-j\omega \tau_{K,m}(\theta_m)}
\end{bmatrix}
\end{equation}
where \(\theta_m\) is the angle of arrival, and \(\omega\) is the angular frequency of the signal. Notice we have adopted a far-field assumption in this signal model, it can be easily extended to include signal model with near-field effect for ultra-short range radars.
Finally, The output of the element can be modeled as:
\begin{equation}
\mathbf{y}(t) = \sum_{m=1}^M \mathbf{A}_m(\theta_m, \alpha_m) \mathbf{s}_m(t) + \mathbf{n}(t)    
\end{equation}
where:
\begin{itemize}
    \item $\mathbf{y}(t) = [x_1(t), x_2(t), \ldots, x_K(t)]^T$
    \item $\mathbf{A}(\theta)$ is a matrix with columns as steering vectors for different angles.
    \item $\mathbf{s}(t)$ represents the source signals at time $t$.
    \item $\mathbf{n}(t)$ represents the noise and interference vector on the arrays.
\end{itemize}

This approach generates ground truth and data in one shot.
The efficient representation further allows to generate training and evaluation data on the fly.
The unlimited data supply in such a setting effectively eliminates the performance gap between training and validation set, and ultimately leads to excellent generalization performance on the data from real radar sensors.
Empirically this is also critical for the zero-shot transfer of the model trained with synthetic data to real data.

\begin{figure}[t]
\centering
\includegraphics[width=\columnwidth]{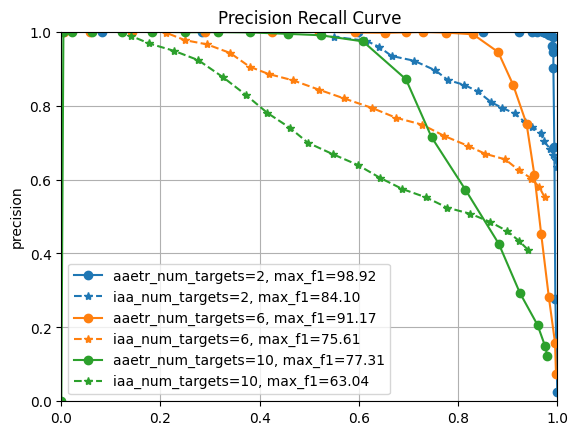} 
\caption{The comparison of PR curves from AAETR and IAA when SNR is set to 35 dB. 
The precision-recall curves reveal how well each algorithm maintains detection performance across varying numbers of targets and dynamic ranges. The AATER generally shows better precision and recall trade off across wide range of threshold values. 
}
\label{fig: PR curves}
\end{figure}

\begin{figure}[t]
\centering
\includegraphics[width=\columnwidth]{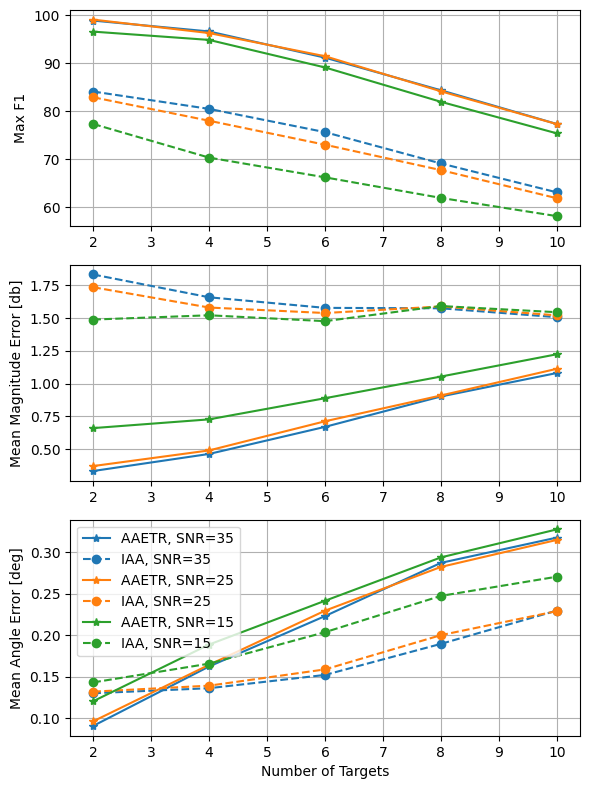} 
\caption{The comparison of the max F1 score and the mean error in magnitude and angle between AAETR and IAA across the number of targets per RD bin. The maximum F1 highlights the scalability of the AATER algorithm. The AATER maintains a significantly higher F1 score across all target numbers and SNRs. AAETR also generates more accurate magnitude prediction compared to IAA. Both methods are able to perform angle prediction with high accuracy. Overall, the robustness of the AAETR to varying levels of SNR is noticeably better than IAA.}
\label{fig: Regression Error}
\end{figure}

\begin{figure*}[t]
\centering
\includegraphics[width=0.9\textwidth]{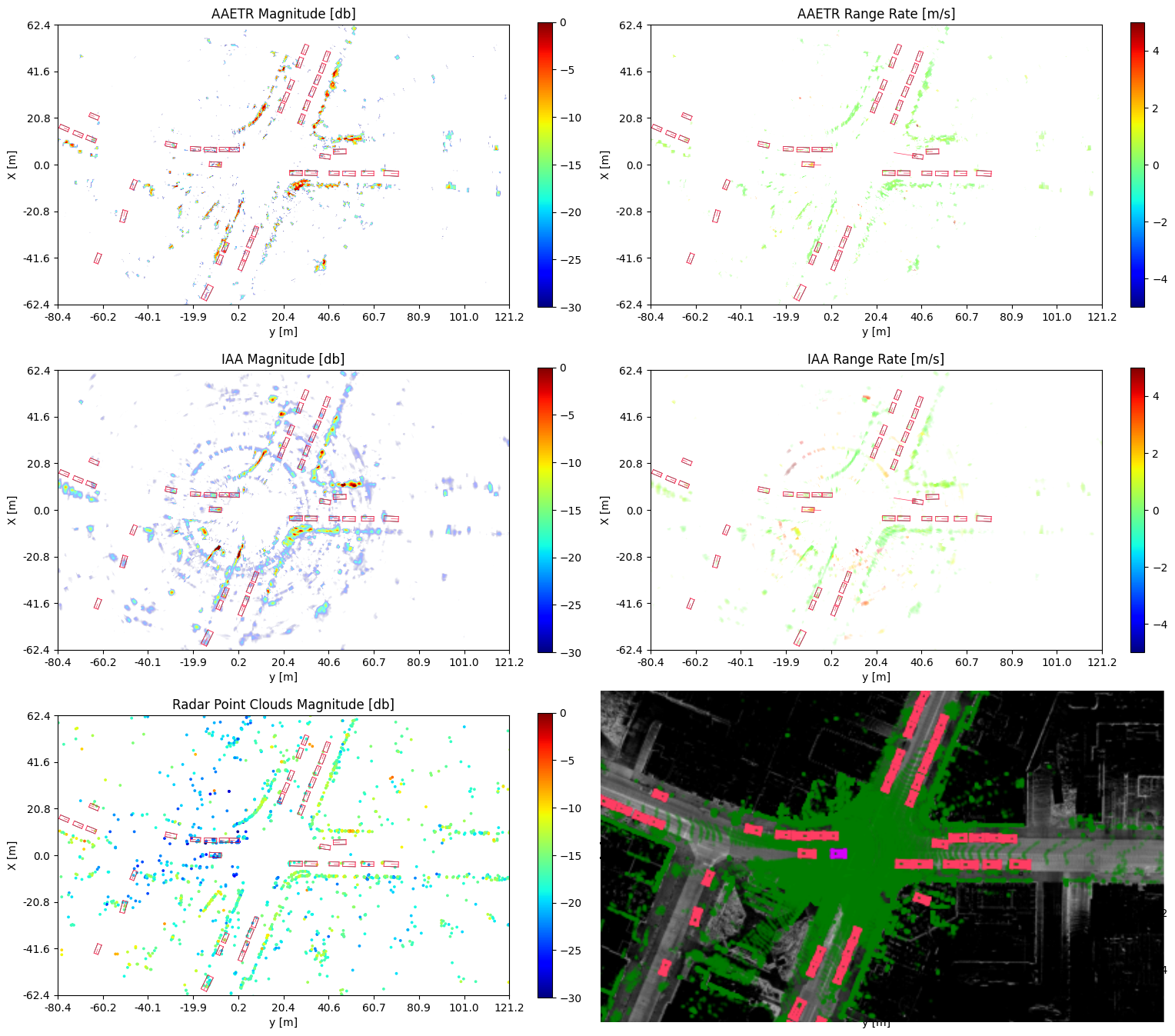} 
\caption{Sample bird's-eye-view rendering of AAETR, IAA, the radar detections from embedded DSP lidar point clouds (bottom right).
Here, the positive x and y directions corresponde to the front and the left of the ego vehicle, respectively.}
\label{fig: sample rendering 1}
\end{figure*}

\begin{figure*}[t]
\centering
\includegraphics[width=0.99\textwidth]{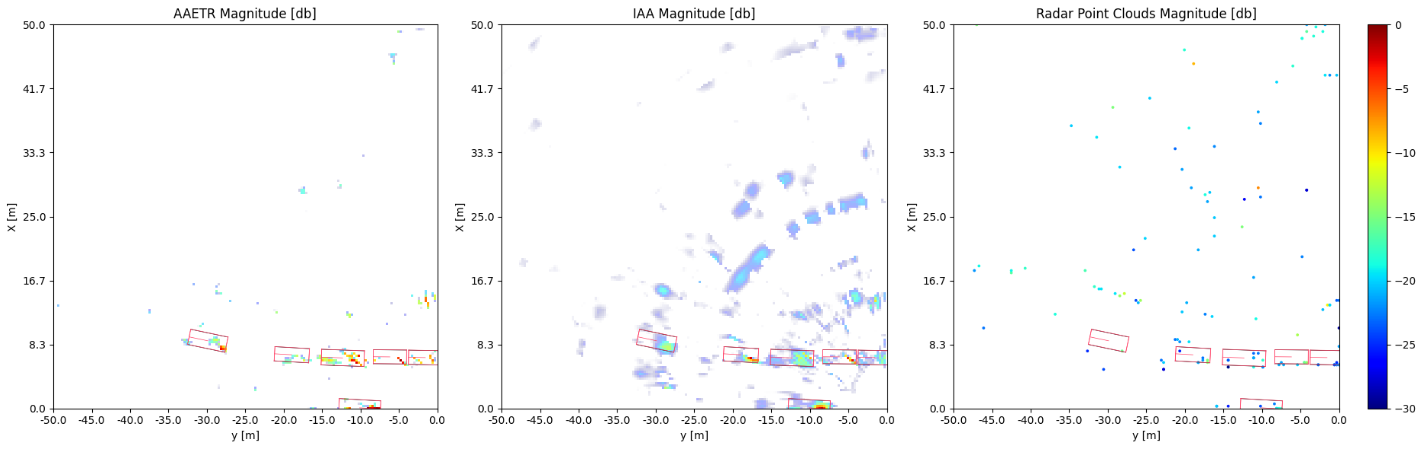} 
\caption{Zoom-in views of the BEV feature maps from AAETR, IAA and radar detections from embedded DSP on the rear left side of the ego vehicle. As indicated by Figure \ref{fig: rear left camera view} and the lidar signal in Figure \ref{fig: sample rendering 1}, the radar signal is supposed to be blocked by the buildings by the street, and any detection should be considered as FP. Therefore, AAETR clearly demonstrates much better precision performance compared to IAA and the radar point clouds.} 
\label{fig: rear left radar view}
\end{figure*}

\begin{figure}[t]
\centering
\includegraphics[width=0.99\columnwidth]{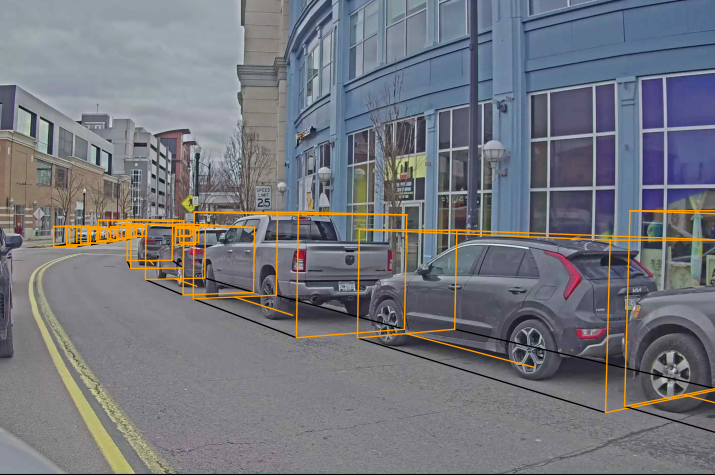} 
\caption{The camera view of the rear left side of the ego vehicle in the scene in Figure \ref{fig: sample rendering 1}.}
\label{fig: rear left camera view}
\end{figure}

\subsection{Training Details}
The model architecture comprises 6 encoder and 6 decoder blocks, each containing layer normalization, dropout, and attention modules. 
This design closely follows the architecture proposed in DETR \cite{carion2020end}, with one key difference: we use learned parameters for both queries and query positions in the decoder, as opposed to zero-initialized queries and learnable query positions in the original DETR implementation. 
The embedding dimension is set to 128, with 80 learnable queries.
In total, the model contains 1.6M parameters.
Training is conducted using a batch size of 32,768 across 16 Nvidia A10 GPUs, with 63M randomly generated samples used throughout the process. 

\subsection{Results}

To evaluate performance, we compare AAETR against IAA under various SNR conditions with multiple target numbers.
A true positive detection is defined as having a predicted probability higher than the confidence threshold and a predicted angle within 0.5 degrees of any ground truth target.
Each evaluation condition uses 4,000 samples from the synthetic dataset, applied consistently for both AAETR and IAA assessments.

Figure \ref{fig: PR curves} presents precision-recall curves for AAETR and IAA at 35 dB SNR with varying target numbers and a dynamic range of 13 dB. The results demonstrate that AAETR significantly outperforms IAA in detection performance, particularly as the number of targets increases. 
Figure \ref{fig: Regression Error} compares the max F1 score and regression errors in detection magnitude and angle across various conditions. 
Although training was conducted at a constant 35 dB SNR, we evaluated the models using data generated at different SNR levels.
Notably, the proposed AAETR method demonstrates robustness to mismatched training and evaluation noise levels.
Moreover, the performance gap between AAETR and IAA widens as noise levels increase.

\subsection{Zero Shot on Real Data}

Beyond its superior metric performance on synthetic data compared to IAA, AAETR also demonstrates excellent zero-shot capability on real data collected by in-house surrounding-view radar systems in dynamic and complex driving scenarios. 
Figure \ref{fig: sample rendering 1} compares AAETR, IAA, and the radar detections from embedded Digital Signal Processing (DSP) module in an urban scene captured by our testing vehicle equipped with multiple radars for surrounding-view perception tasks in Pittsburgh, PA.
The AAETR model, trained on a synthetic dataset generated with in-house radar arrays, is integrated into the radar signal processing chain for RD to RAD conversion.
As AAETR is inherently gridless, summation splatting \cite{niklaus2020softmax} is employed to convert prediction targets to RAD data cubes. 
Each radar is processed independently, with the resulting RAD features resampled into Bird's Eye View (BEV) space using bi-linear interpolation.
For BEV grids covered by multiple radars, maximal magnitude values are used to create surrounding-view BEV features.

Despite being trained solely on synthetic data, AAETR produces higher quality BEV features compared to IAA and the radar detections from embedded DSP. While quantitative evaluation is challenging due to the lack of ground truth in these scenarios, AAETR appears to capture a more comprehensive range of dynamic objects (e.g., moving vehicles) and static features (e.g., street buildings and poles). Moreover, thanks to effective side lobe suppression, AAETR's BEV features exhibit higher precision, as corroborated by lidar point clouds shown in Figure \ref{fig: sample rendering 1}'s bottom right plot.

Two zoom-in views further illustrate AAETR's superior performance.
Figures \ref{fig: front left radar view} and \ref{fig: front left camera view} highlight the front left side of the ego vehicle.
With ground truth vehicle bounding boxes overlaid on the zoomed-in BEV features, both AAETR and IAA provide more semantic information (e.g., width, length) for bounding box detection compared to radar detections from embedded DSP, which often detect only one or two points on distant vehicles.
Notably, AAETR precisely detects a pole at $(x, y) \approx (10, 20)$ with a strong signal, while IAA's signals near the pole are nearly indistinguishable from false positives due to side lobe effects, and radar detections from embedded DSP are sparse in this area.

Figures \ref{fig: rear left radar view} and \ref{fig: rear left camera view} focus on the rear left view of the scene.
As shown in Figure \ref{fig: rear left camera view}, there should be no detections beyond the buildings at the rear left of the ego vehicle. While both IAA and on-device point clouds produce numerous false positive signals beyond the buildings, AAETR generates a clean feature map in this area.
In summary, AAETR's clear separation between positive and negative signals makes it a superior candidate for downstream perception tasks.

\section{Conclusion}
In this paper, we presented AAETR (\textbf{A}ngle of \textbf{A}rrival \textbf{E}stimation with \textbf{TR}ansformer), a deep learning-based gridless AOA algorithm for automotive MIMO radars.
Leveraging a computationally efficient and high-fidelity radar simulation pipeline, AAETR consistently outperforms the classic Iterative Adaptive Approach (IAA) across various conditions, including different numbers of targets and signal-to-noise ratio (SNR) levels.
AAETR's architecture, powered by transformers and trained on an extensive simulated dataset, enables efficient and scalable sparse angle-finding capabilities.
This approach overcomes the limitations of grid-based methods like IAA, such as high computational costs and straddling loss in SNR.
Moreover, AAETR requires fewer tunable hyperparameters and is end-to-end trainable, making it well-suited for integration into deep learning radar perception pipelines.
A key strength of AAETR is its remarkable zero-shot sim-to-real transferability, demonstrated by its strong performance on data collected from testing vehicles in realistic driving scenarios. In these real-world applications, AAETR exhibits superior side lobe suppression compared to IAA while delivering similarly rich information. Additionally, AAETR provides more comprehensive semantic information than radar detections from embedded DSP while maintaining higher precision.
These advantages position AAETR as a promising drop-in module for practical automotive radar systems, potentially enhancing the perception capabilities of ADAS and AV. 

\bibliography{aaai25}

% \appendix
% \section{Reproducibility Checklist}
% \subsection{Theoretical Contribution}
% The methodology and the modeling approach are stated in the section named ``AAETR". 
% \subsection{Datasets}
% The datasets used in this study are synthetically generated. 
% The sensor information and the data generation process is clearly stated in the subsection of ``Synthetic Dataset" in the section of ``Experiments".
% The sizes of the datasets are stated in the subsection of ``Training Details" in the section of ``Experiments".
% \subsection{Computational Experiments}
% Due to the company policy, we cannot open source our model. The model implementation, however, largely follows DETR and their open source implementation \cite{carion2020end}. The difference are stated in the ``AAETR" section. The computational resources required to carry out the experiments are stated in the subsection of `Training Details" in the section of ``Experiments".
% For model optimization, we used `AdamW` with a learning rate of $10^{-3}$.

\end{document}